\documentclass[11pt]{article}

\usepackage[preprint]{acl}

\usepackage{times}
\usepackage{latexsym}
\usepackage{amsmath}
\usepackage{amsmath, amssymb, bm}
\usepackage{booktabs}
\usepackage{graphicx}
\usepackage{enumitem}
\usepackage{booktabs}
\usepackage{multirow}
\usepackage[table]{xcolor}
\usepackage{dblfloatfix} 
\usepackage[T1]{fontenc}

\usepackage[utf8]{inputenc}

\usepackage{microtype}

\usepackage{inconsolata}
\usepackage{fontawesome} 
\usepackage{graphicx}
\title{\textsc{MultiHaluDet}:\\
Multilingual Hallucination Detection via LLM Hidden State Probing}

\author{
 \textbf{Riasad Alvi\textsuperscript{1}},
 \textbf{Nurul Labib Sayeedi\textsuperscript{1}},
 \textbf{Md. Faiyaz Abdullah Sayeedi\textsuperscript{1,2}}
\\
 \textsuperscript{1}United International University,
 \textsuperscript{2}BRAC University
\\
 \small{ralvi212069@bscse.uiu.ac.bd, nsayeedi2410045@bsds.uiu.ac.bd,
   msayeedi212049@bscse.uiu.ac.bd} \\
\small{
   \href{https://github.com/alvi-uiu/MULTIHALUDET}{\faGithub\enspace https://github.com/alvi-uiu/MultiHaluDet}
 }
}

\begin{document}
\maketitle
\begin{abstract}
Hallucinations in Large Language Models (LLMs) represent a critical barrier to their reliable deployment, a vulnerability heavily exacerbated in non-English and resource-constrained contexts. Existing detection approaches that rely on output confidence heuristics or single-layer internal representations frequently fail to capture deep, complex factual inconsistencies across diverse languages. To address this, we introduce \textsc{MultiHaluDet}, a novel four-stage framework that detects multilingual hallucinations by probing the full hidden state trajectories of frozen LLMs without requiring language-specific fine-tuning. Our method extracts sequential features across multiple layers and processes them via a hybrid architecture using multi-scale attention and self-attention pooling. By generating out-of-fold embeddings that feed into a learned ensemble meta-learner, \textsc{MultiHaluDet} captures both fine-grained and coarse-grained patterns of factual inconsistency. Extensive experiments demonstrate that our framework achieves state-of-the-art detection performance, reaching up to 98.55\% AUROC on the English HaluEval and TriviaQA benchmarks using Mistral-7B and LLaMA2-7B architectures. Crucially, we rigorously evaluate our framework's cross-lingual generalization across high (French), medium (Bangla), and low-resource (Amharic) languages. \textsc{MultiHaluDet} demonstrates exceptional representational robustness, consistently outperforming baselines and successfully transferring hallucination detection capabilities across typologically diverse linguistic tiers.
\end{abstract}

\section{Introduction}

Hallucinations in LLMs have emerged as a critical barrier to their reliable deployment across high-stakes domains \citep{farquhar2024detecting, mishra2024fine, varshney2023stitch}. These generations pose significant risks \citep{farquhar2024detecting}, motivating diverse detection techniques. Evidence-based methods \citep{chern2023factool, zhang2024knowhalu} retrieve external information to verify factual consistency, but are computationally intensive. Evidence-free methods leverage inherent model characteristics: logit-based and consistency-based methods estimate uncertainty and output stability, while classification-based methods \citep{orgad2024llms, binkowski2025hallucination} probe internal hidden states without external retrieval.

While internal state probing shows promise \citep{orgad2024llms, binkowski2025hallucination}, current approaches remain inadequate, particularly across diverse languages. Recent work reveals truthfulness information is concentrated in specific tokens \citep{orgad2024llms}; however, single-position methods struggle when non-factual tokens are distributed across sequences. Furthermore, attention-based approaches \citep{chuang2024lookback, binkowski2025hallucination} using simple ratios achieve limited discrimination. Probabilistic frameworks \citep{hou2025probabilistic} require complex reasoning pipelines, while active validation \citep{varshney2023stitch} and tool-augmented systems \citep{chern2023factool, zhang2024knowhalu} introduce substantial latency. As noted in \citet{farquhar2024detecting}, hallucinations manifest as semantic-level confabulations rather than token-level uncertainty. Existing methods largely fail to address these confabulations across non-English and resource-constrained languages.

Transformer architectures process information through deep layers, embedding language-agnostic hallucination signals within hidden state trajectories. Deep sequence modeling with multi-scale feature aggregation offers a promising solution. We introduce \textsc{MultiHaluDet}, a supervised framework leveraging multi-scale attention and transformer encoders to model hidden state dynamics across the full depth of frozen LLMs. Unlike methods focused on individual tokens or static layers, our approach aggregates information across multiple scales through self-attention pooling. We evaluate across multiple architectures and language tiers, demonstrating strong cross-lingual performance. 

Our contributions are summarized as follows:

\begin{itemize}
    \item We introduce \textsc{MultiHaluDet}, a four-stage framework comprising dynamic feature extraction, multi-scale attention encoding, out-of-fold deep feature generation, and learned ensemble meta-learning, which jointly capture fine-grained and coarse-grained patterns from hidden-state trajectories.
    \item We evaluate our framework on the HaluEval and TriviaQA benchmarks using Llama-2-7B and Mistral-7B-Instruct, demonstrating substantial performance gains and representational robustness over existing baselines.
    \item We comprehensively evaluate the cross-lingual generalization of our framework by translating standard benchmarks into French (high-resource), Bangla (medium-resource), and Amharic (low-resource). Our results demonstrate that internal state probing can maintain strong detection signals across typologically diverse linguistic tiers under controlled translation-based evaluation, without language-specific fine-tuning.
\end{itemize}

\section{Related Work}
\label{sec:related}

Hallucination detection approaches fall into three categories: (i) evidence-based methods verifying outputs against external knowledge, (ii) self-detection methods leveraging internal model states, and (iii) consistency-based approaches assessing output stability.

Evidence-based methods retrieve external information for verification. \citet{kale2025lie} converts LLM responses into knowledge graphs for atomic fact verification. \citet{vangala2025hallumat} introduced multi-source retrieval with contradiction graph analysis. \citet{zhang2025hallucination} employed HHEM for lightweight consistency assessment. While effective, these methods depend on retrieval quality, making them vulnerable to gaps and latency.

Self-detection methods probe LLM internal representations without external resources. \citet{su2024unsupervised} introduced MIND, training classifiers on auto-generated pseudo-labels. \citet{liang2025neural} employed Bayesian optimization to identify optimal layer insertion points. \citet{zhang2025detecting} proposed MHAD, selecting specific neurons via linear probing. \citet{kossen2024semantic} approximated semantic uncertainty from hidden states. \citet{chen2024inside} proposed INSIDE, measuring consistency through eigenvalues of response covariances. \citet{kim2025detecting} analyzed layer-wise usable information across transformer depths. \citet{quevedo2024detecting} demonstrated strong detection using only four token probability features. Consistency-based methods assess output stability. \citet{yang2025hallucination} proposed MetaQA, leveraging metamorphic relations for semantic consistency verification.

Despite these advances, challenges remain. Evidence-based methods suffer from retrieval dependency. Self-detection methods relying on single-position representations struggle when non-factual tokens appear at sequence beginnings. Consistency-based methods incur costs from multiple generation passes. \citet{simhi2025trust} demonstrated that LLMs can hallucinate with high certainty despite possessing correct knowledge. \citet{yang2025heaven} revealed the need to distinguish between intelligent and defective hallucinations.

Our work aligns with internal state probing methods but addresses their key limitations. While \citet{liang2025neural} and \citet{zhang2025detecting} focus on specific layers or tokens, we employ dynamic layer sampling and multi-scale attention to aggregate information across the full depth trajectory. Unlike \citet{kossen2024semantic} and \citet{chen2024inside}, which probe specific positions, our architecture processes sequential layer features with transformer encoders and self-attention pooling, enabling adaptive depth selection. Our out-of-fold stacking with ensemble meta-learning provides more robust generalization than single-classifier approaches \citep{su2024unsupervised, liang2025neural}.

\begin{figure*}[t]
    \centering
    \includegraphics[width=\textwidth]{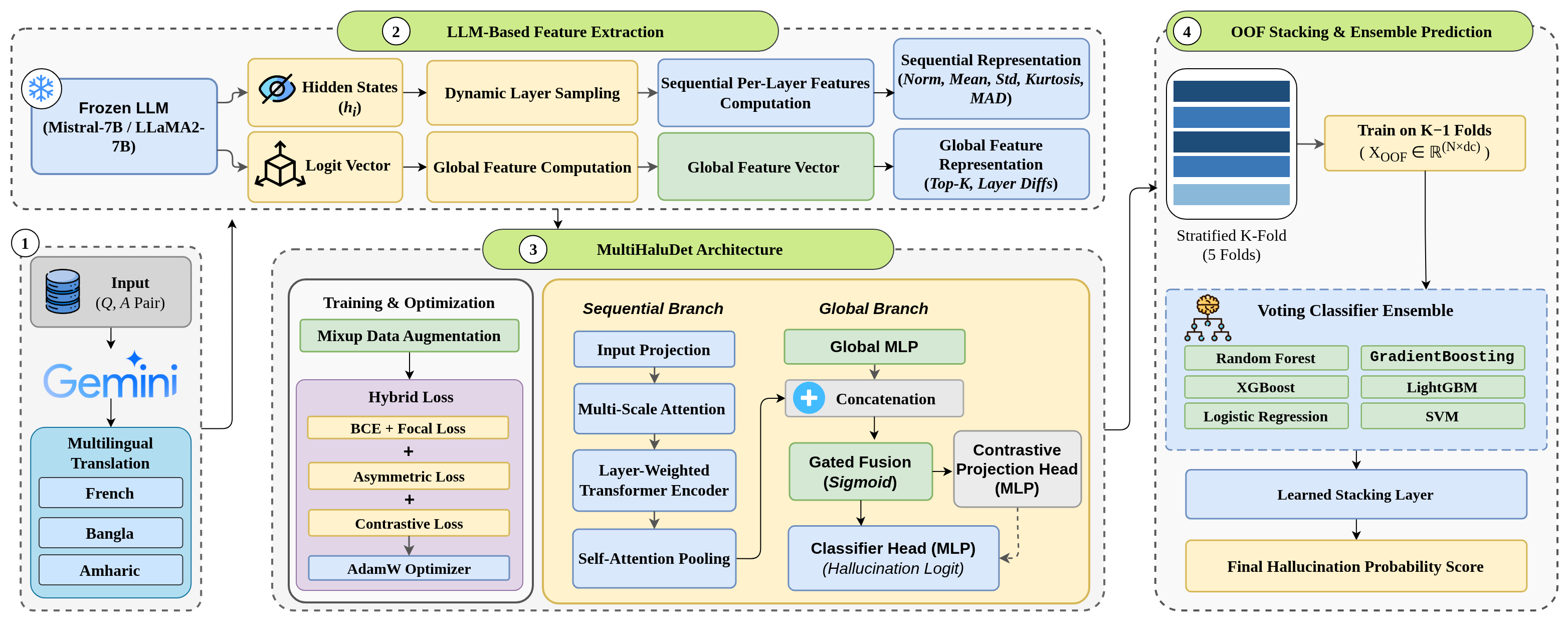}
    \caption{Overview of the four-stage \textsc{MultiHaluDet} framework for multilingual hallucination detection.}
    \label{fig:method}
\end{figure*}

\section{Methodology}
\label{sec:methodology}

We present \textsc{MultiHaluDet} (Figure \ref{fig:method}), a four-stage framework for hallucination detection.

\subsection{LLM-Based Feature Extraction}
\label{sec:features}

\subsubsection{Prompt Construction and Forward Pass}

Let $\mathcal{D} = \{(q_i, a_i, y_i)\}_{i=1}^{N}$ denote a dataset
of question--answer pairs, where $q_i$ is a natural-language question,
$a_i$ is a candidate answer, and $y_i \in \{0,1\}$ is a binary label
indicating whether $a_i$ is a hallucination ($y_i=1$) or a faithful
response ($y_i=0$). Each sample $(q_i, a_i)$ is formatted as a structured natural-language
prompt and tokenized with a fixed maximum sequence length. The prompt
is passed through a frozen, quantized LLM in a single forward pass, yielding a sequence of hidden state tensors $\{\mathbf{H}^{(l)}\}_{l=0}^{L}$, where $L$ is
the number of transformer layers and
$\mathbf{H}^{(l)} \in \mathbb{R}^{T \times d}$ collects the
$d$-dimensional representations of $T$ tokens at layer $l$. The
next-token logit vector $\mathbf{z} \in \mathbb{R}^{V}$ at the final
position is retained for global statistics. No gradient is computed
during feature extraction; the LLM parameters remain fully frozen
throughout all experiments.

\subsubsection{Dynamic Layer Sampling}
\label{sec:layer_sampling}

To ensure architectural compatibility across LLMs of varying depth, we
introduce a \emph{dynamic layer sampling} strategy that maps any
model's $L$ transformer layers to a fixed target count $K$, producing
a uniform sequential representation regardless of model size. Three
regimes are handled without model-specific configuration:

\begin{itemize}[leftmargin=*,nosep]
  \item \textbf{Exact match} ($L = K$): identity mapping
        $\mathcal{I}_{K} = \{1,\ldots,K\}$.
  \item \textbf{Shallow model} ($L < K$): all $L$ layers are used
        and the deepest layer is repeated to pad the sequence to $K$.
  \item \textbf{Deep model} ($L > K$): $K$ indices are selected by
        uniform interpolation over the full depth:
\end{itemize}

\begin{equation}
  \mathcal{I}_{K} = \left\{
    \operatorname{clip}\!\left(
      \left\lfloor 1 + (L-1)\cdot\frac{k}{K-1} \right\rceil,
      1, L
    \right)
  \right\}_{k=0}^{K-1}
  \label{eq:layer_sample}
\end{equation}

\noindent where $\lfloor\cdot\rceil$ denotes rounding to the nearest
integer and $\operatorname{clip}(\cdot,1,L)$ clamps indices to valid
range. This formulation is differentiable with respect to $K$ and
requires no architecture-specific configuration, making it directly
applicable to any transformer-based LLM.

\subsubsection{Sequential Per-Layer Features}

For each sampled layer index $l \in \mathcal{I}_{K}$, we extract a
compact descriptor from two complementary views of the hidden state
tensor: the last-token representation
$\mathbf{h}^{(l)} = \mathbf{H}^{(l)}_{T,:}$, which reflects the
model's final contextual state, and the sequence mean
$\bar{\mathbf{h}}^{(l)} = \frac{1}{T}\sum_{t=1}^{T}\mathbf{H}^{(l)}_{t,:}$,
which captures the aggregate token-level context. From these, we
compute a per-layer descriptor
$\mathbf{s}^{(l)} \in \mathbb{R}^{d_s}$ comprising distributional
statistics: the $\ell_2$ norm, mean, standard deviation, extremal
values, activation sparsity, near-zero mass, and the kurtosis and
median absolute deviation (MAD) of the hidden state, capturing the
peakedness and robust dispersion of the activation distribution:


\begin{equation}
\begin{aligned}
\kappa(\mathbf{h}) &= \frac{1}{d}\sum_{j=1}^{d}\left(\frac{h_j - \bar{h}}{\hat{\sigma}}\right)^{4}, \\
\operatorname{MAD}(\mathbf{h}) &= \operatorname{median}_j\!\left(|h_j - \operatorname{median}(\mathbf{h})|\right)
\end{aligned}
\label{eq:layer_stats}
\end{equation}

\noindent where $\bar{h}$ and $\hat{\sigma}$ denote the empirical mean and
standard deviation of the hidden-state vector, and
$\operatorname{median}(\mathbf{h})$ is its coordinate-wise median.
Collecting descriptors
across all sampled layers yields the sequential representation
$\mathbf{S} \in \mathbb{R}^{K \times d_s}$, which encodes how the
LLM's internal dynamics evolve as a function of depth.

\subsubsection{Anchor-Based Depth Probing}

To enable consistent cross-architecture comparisons at semantically
meaningful network depths, we define four \emph{anchor layers}
corresponding to proportional depth fractions
$\{\alpha_j\}_{j=1}^{4}$. For each anchor, the closest available
sampled layer is identified:

\begin{equation}
  l^*_j = \underset{l \,\in\, \mathcal{I}_K}{\arg\min}
    \left|\, l - \left\lfloor \alpha_j \cdot L \right\rceil \,\right|
  \label{eq:anchor}
\end{equation}

\noindent and the corresponding descriptor $\mathbf{s}^{(l^*_j)}$ is
retained as an anchor feature. This construction ensures that features
at early, middle, and late network stages are explicitly represented
in the global feature vector regardless of the total model depth,
providing a principled substitute for hardcoded layer indices.

\subsubsection{Global Feature Vector}

A global feature vector $\mathbf{g} \in \mathbb{R}^{d_g}$ aggregates
information across the full forward pass. It comprises: (i) the
top-$k$ next-token probabilities and their pairwise differences,
capturing the sharpness of the model's next-token distribution;
(ii) the entropy, standard deviation, and maximum of the logit vector
$\mathbf{z}$; (iii) first- and second-order statistics of the
layer-wise $\ell_2$-norm trajectory
$\{r_l\}_{l \in \mathcal{I}_K}$, $r_l = \|\mathbf{h}^{(l)}\|_2$,
capturing norm growth and volatility across depth; (iv) the anchor
descriptors from Eq.~\ref{eq:anchor}; and (v) cross-feature
interaction terms between logit statistics and norm dynamics, which
encode the coupling between the model's output confidence and its
internal representational geometry. Together, $\mathbf{S}$ and
$\mathbf{g}$ form the complete feature representation passed to classification stage.

\subsection{MultiHaluDet Architecture}
\label{sec:model}

\subsubsection{Projection and Multi-Scale Attention}

The sequential input $\mathbf{S}$ is first projected into a uniform
hidden space of dimension $H$ via a linear layer followed by layer
normalization and a GELU activation. A \emph{multi-scale attention}
module then processes the projected sequence at multiple temporal
resolutions simultaneously. For each scale factor $c \in \mathcal{C}$,
the sequence is locally average-pooled to compress $K$ positions into
$\lceil K/c \rceil$ positions, passed through a scale-specific linear
projection, and upsampled back to length $K$ via nearest-neighbor
interpolation. The contributions of each scale are combined through a
learned position-wise gate:

\begin{equation}
  \tilde{\mathbf{h}}_t =
    \sum_{c \in \mathcal{C}}
    w_{t,c}\;\mathbf{P}_c\!\left(
      \operatorname{Upsample}\!\left(
        \operatorname{Pool}_c\!\left(\mathbf{H}_{\text{proj}}\right)
      \right)_t
    \right)
  \label{eq:multiscale}
\end{equation}

\noindent where $w_{t,c} = \operatorname{softmax}(\mathbf{W}_g
\mathbf{h}_t)_c$ is the position-wise scale gate computed from the
original projected hidden state, and $\mathbf{P}_c$ is the per-scale
linear projection. The output is combined with the original projection
via a residual connection and layer normalization, preserving
fine-grained layer information while enriching it with multi-scale
context.

\subsubsection{Layer-Weighted Transformer Encoder}

Because different LLM layers carry information of varying
discriminative value for hallucination detection, we modulate the
fused sequence by a learnable, softmax-normalized importance vector
$\bm{\lambda} \in \mathbb{R}^{K}$ before encoding:

\begin{equation}
\begin{aligned}
\hat{\mathbf{h}}_k &=
\bar{\lambda}_k \cdot \tilde{\mathbf{h}}_k + \mathbf{p}_k,
\quad k = 1, \ldots, K, \\
\bar{\bm{\lambda}} &= \operatorname{softmax}(\bm{\lambda})
\end{aligned}
\label{eq:layer_weight}
\end{equation}

\noindent where $\mathbf{p}_k$ is a learned positional embedding that
encodes the relative depth of each sampled layer within the
representational sequence. The modulated sequence is encoded by a
stack of Pre-LN Transformer encoder layers with multi-head
self-attention and GELU feed-forward sublayers, enabling the model to
attend across LLM depths and capture long-range inter-layer
dependencies in the hidden state trajectory.

\subsubsection{Self-Attention Pooling}

Rather than discarding positional information through mean pooling or
relying on a fixed summary token, we aggregate the transformer output
via a learned attention pooling mechanism. A two-layer MLP with a
$\tanh$ nonlinearity assigns a scalar relevance score to each position,
and the final sequential representation is their weighted sum:

\begin{equation}
  \mathbf{u} = \sum_{k=1}^{K} \alpha_k \mathbf{e}_k, \quad
  \alpha_k = \frac{\exp\!\left(a(\mathbf{e}_k)\right)}
                  {\sum_{j=1}^{K} \exp\!\left(a(\mathbf{e}_j)\right)}
  \label{eq:attn_pool}
\end{equation}

\noindent where $\mathbf{e}_k$ is the encoder output at position $k$
and $a:\mathbb{R}^H \to \mathbb{R}$ is the learned scoring MLP.
This allows the model to focus on the LLM layers most informative
for the current input, providing a form of input-adaptive depth
selection.

\subsubsection{Global Branch and Gated Fusion}

The global feature vector $\mathbf{g}$ is processed independently
through a two-layer MLP with layer normalization and dropout, yielding
a compact global representation $\mathbf{v} \in \mathbb{R}^{H/2}$.
The sequential representation $\mathbf{u} \in \mathbb{R}^{H}$ and
global representation are concatenated to form the joint embedding
$\mathbf{c} = [\mathbf{u};\mathbf{v}] \in \mathbb{R}^{3H/2}$.
A sigmoid gate then performs element-wise re-weighting to suppress
uninformative dimensions:

\begin{equation}
  \tilde{\mathbf{c}} = \mathbf{c} \odot
    \sigma\!\left(\mathbf{W}_{\text{gate}}\,\mathbf{c}
    + \mathbf{b}_{\text{gate}}\right)
  \label{eq:gate}
\end{equation}

\noindent The gated representation $\tilde{\mathbf{c}}$ is passed
through a three-layer MLP classifier to produce the final
hallucination logit. A separate two-layer projection head maps
$\tilde{\mathbf{c}}$ onto a unit hypersphere for the contrastive
objective described in Section \ref{sec:exp}.

\subsection{Out-of-Fold Stacking}
\label{sec:oof}

To obtain unbiased deep representations for the meta-learner without
data leakage, we employ $K$-fold out-of-fold (OOF) stacking over the
training set. For fold $k$, a fresh \textsc{MultiHaluDet} instance
is trained on the remaining $K-1$ folds and used to extract the gated
fusion representation $\tilde{\mathbf{c}}$ for the held-out fold. After
all $K$ folds, the OOF features form a complete training matrix:

\begin{equation}
  \mathbf{X}^{\text{OOF}} \in \mathbb{R}^{N_{\text{train}} \times d_c},
  \quad d_c = \dim(\tilde{\mathbf{c}})
  \label{eq:oof}
\end{equation}

\noindent with every training sample represented exactly once without
ever being used in its own fold's training. Test-set representations
are obtained by averaging the gated fusion embeddings across all $K$
fold models in feature space:

\begin{equation}
  \tilde{\mathbf{c}}^{\text{test}} =
    \frac{1}{K}\sum_{k=1}^{K} \tilde{\mathbf{c}}^{(k)\text{test}}
  \label{eq:test_avg}
\end{equation}

\noindent This averaging operates at the representation level rather
than the probability level, providing implicit test-time ensembling
and producing a more stable input to the meta-learner. The OOF and
test features are standardized before being passed to the ensemble
stage.

\subsection{Ensemble Meta-Learner}
\label{sec:ensemble}

The OOF deep features serve as input to a diverse ensemble of
classifiers comprising gradient-boosted trees, random forests, a
kernel support vector machine, a multi-layer perceptron, and a
logistic regression baseline. Model diversity across both inductive
biases (linear, kernel, tree, neural) and hyperparameter profiles
reduces variance and improves robustness to the distributional
properties of the deep features. Rather than fixed heuristic weights,
the ensemble prediction is formed by stacking the base classifier
probability estimates through a logistic meta-regressor trained on
held-out out-of-fold predictions. Denoting the log-odds transform of
each base probability by $\ell_m = \log(\hat{p}_m / (1-\hat{p}_m))$,
the final ensemble probability is:

\begin{equation}
  \hat{p}_{\text{ens}} = \sigma\!\left(\beta_0 + \sum_{m=1}^{M} \beta_m \,\ell_m\right)
  \label{eq:ensemble}
\end{equation}

\noindent where $\sigma(\cdot)$ is the sigmoid function and the
coefficients $\{\beta_m\}_{m=0}^{M}$ are learned by an $\ell_2$-regularized
logistic regression meta-learner trained on a secondary hold-out split
of the out-of-fold probability outputs. This formulation adapts the
ensemble composition to the empirical discriminative strength of each
base learner while producing inherently calibrated probability estimates.
All component models are configured with balanced class weights to
maintain sensitivity under any residual distributional skew. The
classification threshold is set by Youden's J statistic
$\tau^* = \arg\max_\tau[\mathrm{TPR}(\tau) - \mathrm{FPR}(\tau)]$,
which maximizes the simultaneous improvement in sensitivity and specificity.

\subsection{Multilingual Adaptation Strategy}

To evaluate the multilingual generalization of our hidden state probing framework across varying resource tiers, we translated the source English datasets into French (high-resource), Bangla (medium-resource), and Amharic (low-resource) utilizing the Gemini 2.5 Flash model. Formally, let the original source dataset be defined as $\mathcal{D}_{en}=\{(q_{i},a_{i},y_{i})\}_{i=1}^{N}$. We introduce a translation mapping $\mathcal{T}_{\ell}$ for the set of target languages $\mathcal{L} = \{\text{fr}, \text{bn}, \text{am}\}$. For every query-response pair, this operation strictly preserves the original binary hallucination label, yielding a new language-specific instance defined as: $(q_{i}^{(\ell)}, a_{i}^{(\ell)}, y_{i}^{(\ell)}) = (\mathcal{T}_{\ell}(q_{i}), \mathcal{T}_{\ell}(a_{i}), y_{i})$. The multilingual evaluation space is then constructed as the union of the original English data and all language-specific subsets.


To guarantee the semantic integrity and context preservation of the translated data, we conducted human evaluation for quality assurance. We sampled 100 instances per language from each dataset, resulting in 300 evaluated samples per dataset and a total of 600 evaluated samples across both datasets. As three of the authors are native Bangla speakers, the Bangla subset was evaluated directly. For French and Amharic, we employed a back-translation methodology, translating the target samples back into English, to verify semantic equivalence against the original source. This manual evaluation confirmed an initial translation accuracy of 96\%. The remaining 4\% of instances, which exhibited minor semantic drift or structural errors, were explicitly polished and regenerated using Gemini 2.5 Flash to ensure strict alignment with the source truth conditions.

By passing these high-fidelity translated inputs through the frozen LLM, we generate the sequential representation $S$ and global feature vector $g$ natively within each target language's token space. This formulation ensures that the subsequent deep architecture evaluates intrinsic, language-agnostic hallucination signals embedded within the model's internal dynamics, allowing us to probe representation alignment without requiring language-specific fine-tuning.

\section{Experimental Setup}
\label{sec:exp}

\subsection{Dataset}

We evaluate on two question-answering benchmarks. Let $\mathcal{Q}$ denote the set of questions and $\mathcal{A}$ the set of candidate answers. Each dataset $\mathcal{D} = \{(q_i, a_i, y_i)\}_{i=1}^{N}$ consists of $N$ samples where $q_i \in \mathcal{Q}$ is a question, $a_i \in \mathcal{A}$ is an answer, and $y_i \in \{0, 1\}$ is a binary label indicating non-hallucination ($y_i=0$) or hallucination ($y_i=1$). HaluEval \citep{li2023halueval} provides $N=10,000$ human-annotated QA pairs with native hallucination labels derived from real LLM outputs, where each $(q_i, a_i, y_i)$ is explicitly labeled based on factual consistency with verified knowledge sources. TriviaQA \citep{joshi2017triviaqa} is adapted for our evaluation by collecting realistic, model-generated hallucinations. Let $\mathcal{D}_{\text{Trivia}} = \{(q_i, a_i^*)\}_{i=1}^{M}$ denote the original dataset, where $q_i$ is a question and $a_i^*$ is the ground-truth correct answer. To generate plausible hard negatives, we prompt an early-generation language model known for its propensity to hallucinate, Gemma-2-2B ($\sim32\% \text{-} 38\%$ hallucination rate), to answer each $q_i$. Responses that are plausible but factually incorrect are extracted and denoted as $a_i^-$. We then construct our final evaluation dataset by pairing each question with its true correct answer and its corresponding model-generated hallucination. To facilitate our multilingual evaluation, we also expanded these source English datasets into three different target languages.  


\subsection{Baselines}

We compare against four types of hallucination detection methods. \textbf{Prompt-based methods} include P(True) \citep{kadavath2022language}, which utilizes a simple prompt template to enable the model to assess the correctness of its own response. \textbf{Logit-based methods} use the uncertainty of LLM outputs to detect hallucination; we adopt AvgProb and AvgEnt from \citet{huang2023look} to aggregate logit-based uncertainty across all tokens, and also compare with EUBHD \citep{su2024unsupervised}, which focuses on key tokens rather than considering all tokens. \textbf{Consistency-based methods} are motivated by the idea that consistent responses indicate factual knowledge; we apply Unigram and NLI variants, as well as INSIDE \citep{chen2024inside}, which leverages eigenvalues of the covariance matrix of responses. \textbf{Classification-based methods} train a classifier on labeled statements; we compare with SAPLMA \citep{azaria2023internal}, which trains a classifier on the last token's last-layer hidden state; MIND \citep{su2024unsupervised}, which uses unsupervised training on auto-generated pseudo-labels; and Probe@Exact, which relies on information from potentially correct tokens. We also include HD-NDEs \citep{li2025hd} (Neural ODEs, CDEs, and SDEs), which model hidden state trajectories using neural differential equations.

\subsection{Hyperparameters}

We extract hidden states from frozen LLMs in half-precision without gradient computation. For all LLM forward passes and synthetic generation tasks, the temperature is set to $0.0$ to ensure deterministic outputs. Dynamic layer sampling maps variable-depth models to $K=32$ uniform indices. The deep architecture uses hidden dimension $d=384$, 8 attention heads, and 6 transformer encoder layers. Training employs AdamW optimizer with learning rate $2\times10^{-4}$, weight decay $6\times10^{-5}$, and ReduceLROnPlateau scheduling over 45 epochs with early stopping patience of 15 epochs. We apply a composite loss combining BCE, focal, asymmetric, and contrastive objectives with label smoothing, along with data augmentation via Mixup and CutMix. We employ 5-fold stratified cross-validation with out-of-fold feature extraction to prevent leakage. The stacked ensemble combines 6 classifiers (RandomForest, XGBoost, GradientBoosting, LightGBM, LogisticRegression, Support Vector Machine). Their probability outputs are fused by a logistic meta-regressor trained on held-out out-of-fold predictions, producing inherently calibrated ensemble probabilities. All experiments are conducted on AMD Ryzen 5 8500G CPU with 32GB RAM and NVIDIA GeForce RTX 5060 Ti GPU with 16GB VRAM.

\subsection{Evaluation Metric}

We utilize AUROC (\%), which stands for the area under the ROC curve, to objectively evaluate the effectiveness of models. The higher the value of AUROC, the stronger the ability of this method for hallucination detection. All experiments employ 5-fold cross-validation with stratified sampling to ensure reliable estimates.


\section{Results and Analysis}


\subsection{RQ1: To what extent does MultiHaluDet improve hallucination detection compared to existing internal-state and confidence based baselines?}
\label{sec:main_results}

Table~\ref{tab:main_results} presents the detection performance (AUROC \%) of \textsc{MultiHaluDet} compared against thirteen baseline methods, evaluated across both the HaluEval and TriviaQA datasets using Mistral-7B and LLaMA2-7B.

\begin{table}[t]
\centering
\tiny
\setlength{\tabcolsep}{5pt}
\renewcommand{\arraystretch}{1.05}
\begin{tabular}{lcc|cc}
\toprule
\rowcolor[HTML]{EFEFEF}
\multirow{2}{*}{\textbf{Method}} & \multicolumn{2}{c}{\textbf{HaluEval}} & \multicolumn{2}{c}{\textbf{TriviaQA}} \\
\cmidrule(lr){2-3} \cmidrule(lr){4-5}
\rowcolor[HTML]{EFEFEF}
 & \textbf{Mistral-7B} & \textbf{LLaMA2-7B} & \textbf{Mistral-7B} & \textbf{LLaMA2-7B} \\
\midrule
P(True)        & 49.7 & 46.7 & 48.3 & 42.3 \\
AvgProb        & 43.6 & 42.1 & 48.5 & 44.1 \\
AvgEnt         & 49.7 & 47.3 & 47.6 & 41.1 \\
EUBHD          & 70.5 & 71.9 & 80.6 & 80.5 \\
Unigram        & 62.3 & 58.2 & 59.5 & 56.8 \\
NLI            & 63.1 & 61.3 & 61.4 & 59.4 \\
INSIDE         & 76.0 & 74.5 & 81.3 & 81.7 \\
SAPLMA         & 89.4 & 87.0 & 84.1 & 80.0 \\
MIND           & 94.5 & 86.1 & 84.5 & 79.4 \\
Probe@Exact    & 93.4 & 88.3 & 84.1 & 81.3 \\
Neural ODEs    & 91.2 & 89.5 & 83.7 & 81.7 \\
Neural CDEs    & 95.4 & 91.4 & 84.1 & 83.7 \\
Neural SDEs    & 93.7 & 92.8 & 85.1 & 81.0 \\
\midrule
\rowcolor[HTML]{E6E0FA}
\textbf{MultiHaluDet} & \textbf{98.43} & \textbf{98.55} & \textbf{98.30} & \textbf{98.26} \\
\bottomrule
\end{tabular}
\caption{Hallucination detection performance (AUROC \%) on HaluEval and TriviaQA using Mistral-7B-Instruct and LLaMA2-7B. Best results in bold.}
\label{tab:main_results}
\end{table}


\textbf{Failure of Surface-Level Heuristics.} The results reveal a clear hierarchy in the efficacy of different detection paradigms. Token-level probability and entropy metrics (P(True), AvgProb, AvgEnt) systematically fail to provide meaningful detection signals, hovering around or below random chance (41.1\% -- 49.7\% AUROC). This confirms that raw output confidence is severely miscalibrated and insufficient for identifying deep factual inconsistencies. Intermediate methods incorporating structural linguistic features (Unigram, NLI) offer only marginal improvements, generally plateauing in the low 60\% range.

\textbf{Representational Baselines.} Approaches that leverage deeper internal representations show significantly more promise. Methods like SAPLMA, MIND, and traditional probing (Probe@Exact) push performance into the 80\%--94\% range. The strongest baselines are those modeling the continuous dynamics of hidden states (Neural ODEs, CDEs, SDEs). Neural CDEs, in particular, achieve a competitive 95.4\% on Mistral-7B for HaluEval. However, these methods exhibit high variance between model architectures; for instance, MIND's performance drops sharply from 94.5\% on Mistral-7B to 86.1\% on LLaMA2-7B.

\textbf{Ours.} \textsc{MultiHaluDet} achieves state-of-the-art performance across all experimental conditions, substantially outperforming the strongest continuous-time baselines. On HaluEval, we achieve 98.43\% AUROC with Mistral-7B and 98.55\% with LLaMA2-7B. Crucially, \textsc{MultiHaluDet} demonstrates exceptional cross-architecture robustness. While nearly all baselines suffer noticeable performance degradation when transitioning from Mistral to LLaMA2, our framework maintains near-identical, near-perfect efficacy (actually scoring slightly higher on LLaMA2 in HaluEval). Similarly, on the TriviaQA dataset, which features plausible but factually incorrect synthetic hard negatives, we achieve 98.30\% and 98.26\% respectively. These consistent gains across diverse datasets and architectures demonstrate that our multi-scale attention mechanism and out-of-fold stacking approach successfully aggregate robust, architecture-agnostic signals from LLM hidden state trajectories.

\subsection{RQ2: How robust is MultiHaluDet across typologically diverse languages with varying resource availability?}
\label{sec:multilingual}
To demonstrate the cross-lingual generalization of \textsc{MultiHaluDet}, we extend our evaluation beyond English to include three typologically diverse languages: French, Bangla, and Amharic. 

\begin{table}[t]
\centering
\tiny
\setlength{\tabcolsep}{5pt}
\begin{tabular}{llcc}
\toprule
\rowcolor[HTML]{EFEFEF}
\textbf{Language} & \textbf{Method} & \textbf{HaluEval} (\textcircled{\tiny M} / \textcircled{\tiny L}) & \textbf{TriviaQA} (\textcircled{\tiny M} / \textcircled{\tiny L}) \\
\midrule
\multirow{2}{*}{English (Base)}
& Best Baseline  & 95.4 / 92.8 & 85.1 / 83.7 \\
& \textbf{MultiHaluDet} & \cellcolor[HTML]{E6E0FA}\textbf{98.4} / \cellcolor[HTML]{E6E0FA}\textbf{98.5} & \cellcolor[HTML]{E6E0FA}\textbf{98.3} / \cellcolor[HTML]{E6E0FA}\textbf{98.2} \\
\midrule
\multirow{2}{*}{French (High)}
& Best Baseline  & 92.1 / 89.4 & 81.3 / 80.5 \\
& \textbf{MultiHaluDet} & \cellcolor[HTML]{E6E0FA}\textbf{96.2} / \cellcolor[HTML]{E6E0FA}\textbf{95.8} & \cellcolor[HTML]{E6E0FA}\textbf{95.5} / \cellcolor[HTML]{E6E0FA}\textbf{94.9} \\
\midrule
\multirow{2}{*}{Bangla (Medium)}
& Best Baseline  & 78.4 / 75.1 & 69.2 / 67.8 \\
& \textbf{MultiHaluDet} & \cellcolor[HTML]{E6E0FA}\textbf{89.1} / \cellcolor[HTML]{E6E0FA}\textbf{88.4} & \cellcolor[HTML]{E6E0FA}\textbf{87.6} / \cellcolor[HTML]{E6E0FA}\textbf{86.3} \\
\midrule
\multirow{2}{*}{Amharic (Low)}
& Best Baseline  & 62.3 / 59.8 & 54.1 / 52.6 \\
& \textbf{MultiHaluDet} & \cellcolor[HTML]{E6E0FA}\textbf{78.5} / \cellcolor[HTML]{E6E0FA}\textbf{76.2} & \cellcolor[HTML]{E6E0FA}\textbf{75.8} / \cellcolor[HTML]{E6E0FA}\textbf{73.4} \\
\bottomrule
\multicolumn{4}{l}{\vspace{-2mm}} \\ 
\multicolumn{4}{l}{\scriptsize \textcircled{\tiny M}: Mistral-7B \quad \textcircled{\tiny L}: LLaMA2-7B} \\
\end{tabular}
\caption{Cross-lingual hallucination detection performance. MultiHaluDet consistently outperforms the best baseline across all resource settings.}
\label{tab:multilingual}
\end{table}

Table \ref{tab:multilingual} shows that hallucination detection performance correlates with the representational frequency of these languages in the base models. \textsc{MultiHaluDet} maintains exceptionally strong performance on French, achieving 96.2\% and 95.8\% on HaluEval for Mistral-7B and LLaMA2-7B respectively, trailing the English baselines (98.4\% and 98.5\%) by only a marginal fraction. Similar high retention is observed on TriviaQA, where French scores reach 95.5\% and 94.9\%. For Bangla, we observe a more noticeable but controlled degradation, yielding HaluEval scores of 89.1\% (Mistral-7B) and 88.4\% (LLaMA2-7B), alongside 87.6\% and 86.3\% on TriviaQA. While the framework successfully adapts to the language, accurately detecting factual inconsistencies requires the model to effectively navigate Bangla's rich morphology and the complex linguistic adaptations necessary for regional dialect variations. 

Amharic, a severely low-resource language with limited representation in both Mistral and LLaMA2, presents the greatest challenge. The base models inherently struggle with factual recall in Amharic, which limits the quality of the internal representations our detector relies on. As a result, absolute performance understandably drops to 78.5\% and 76.2\% on HaluEval, and 75.8\% and 73.4\% on TriviaQA for Mistral-7B and LLaMA2-7B, respectively. However, \textsc{MultiHaluDet} still manages to extract meaningful detection signals, maintaining a robust lead over random chance and standard confidence baselines.

\subsection{Ablation Study}
\label{sec:ablation}

To understand the necessity of our proposed architectural choices, we conduct an ablation study by systematically removing key components of the \textsc{MultiHaluDet} framework. We evaluate the degraded models on both HaluEval and TriviaQA using Mistral-7B and LLaMA2-7B. The results are summarized in Table~\ref{tab:ablation}.

\begin{table}[t]
\centering
\scriptsize
\setlength{\tabcolsep}{4pt}
\begin{tabular}{lcc|cc}
\toprule
\rowcolor[HTML]{EFEFEF}
\multirow{2}{*}{\textbf{Method}} & \multicolumn{2}{c}{\textbf{Mistral-7B}} & \multicolumn{2}{c}{\textbf{LLaMA2-7B}} \\
\cmidrule(lr){2-3} \cmidrule(lr){4-5}
\rowcolor[HTML]{EFEFEF}
 & \textbf{HaluEval} & \textbf{TriviaQA} & \textbf{HaluEval} & \textbf{TriviaQA} \\
\midrule
\textbf{Full} & \textbf{\textcolor{green!60!black}{98.43}} & \textbf{\textcolor{green!60!black}{98.30}} & \textbf{\textcolor{green!60!black}{98.55}} & \textbf{\textcolor{green!60!black}{98.26}} \\
\midrule
w/o MSA & 91.45 \textcolor{red}{\tiny ($\downarrow$6.98)} & 90.82 \textcolor{red}{\tiny ($\downarrow$7.48)} & 92.14 \textcolor{red}{\tiny ($\downarrow$6.41)} & 91.33 \textcolor{red}{\tiny ($\downarrow$6.93)} \\
w/o OOF & 88.67 \textcolor{red}{\tiny ($\downarrow$9.76)} & 87.41 \textcolor{red}{\tiny ($\downarrow$10.89)} & 89.25 \textcolor{red}{\tiny ($\downarrow$9.30)} & 88.19 \textcolor{red}{\tiny ($\downarrow$10.07)} \\
w/o TP  & 93.28 \textcolor{red}{\tiny ($\downarrow$5.15)} & 92.56 \textcolor{red}{\tiny ($\downarrow$5.74)} & 93.71 \textcolor{red}{\tiny ($\downarrow$4.84)} & 93.04 \textcolor{red}{\tiny ($\downarrow$5.22)} \\
\bottomrule
\end{tabular}
\caption{Ablation study of \textsc{MultiHaluDet}. Performance is measured in AUROC (\%). The red text indicates the absolute performance drop when a specific core component is removed, confirming the structural necessity of the complete framework. Abbreviations -- w/o: without, MSA: Multi-Scale Attention, OOF: Out-of-Fold Stacking, TP: Trajectory Probing.}
\label{tab:ablation}
\end{table}

The removal of the Out-of-Fold (OOF) Stacking mechanism causes the most severe degradation across all configurations, resulting in a precipitous drop of nearly 10 percentage points (falling to 88.67\% on Mistral-7B HaluEval and 87.41\% on TriviaQA). Without OOF stacking, the meta-classifier heavily overfits to the localized noise of early hidden layers, completely failing to generalize when faced with the plausible hard negatives in TriviaQA. This confirms that our stacking approach is non-negotiable for robust feature aggregation.

Bypassing the Multi-Scale Attention module and relying on standard global average pooling results in a substantial 6--8\% drop in AUROC. Hallucination signals are not uniformly distributed across the generation trajectory; they often manifest as sudden, localized semantic shifts in the middle layers. The sharp decline in performance without this module proves that capturing these local trajectory shifts is strictly necessary, and standard pooling mechanisms are too coarse to detect subtle factual deviations.

Finally, replacing our continuous trajectory probing with a static, final-layer representation (w/o Trajectory Probing) degrades performance by approximately 5 percentage points. While the final layer contains significant semantic information, this result empirically proves our core hypothesis: the \textit{process} of how a model arrives at an answer (the hidden state evolution) contains critical truthfulness signals that are permanently lost if one only analyzes the final output state.

\section{Conclusion}
We presented \textsc{MultiHaluDet}, a four-stage framework that detects hallucinations by probing the hidden state trajectories of frozen LLMs without fine-tuning. By integrating multi-scale attention with out-of-fold deep feature generation and learned ensemble meta-learning, our approach effectively captures complex semantic shifts indicating factual inconsistencies. Extensive evaluations show that \textsc{MultiHaluDet} achieves state-of-the-art performance on standard benchmarks, while also demonstrating robust cross-lingual generalization across high, medium, and low-resource languages.

\newpage
\clearpage

\section*{Limitations}

While \textsc{MultiHaluDet} demonstrates strong performance in detecting multilingual hallucinations, several limitations must be acknowledged. 

First, our framework inherently relies on white-box access to the internal hidden states and logits of the target LLMs. Consequently, it cannot be directly applied to proprietary, black-box models (e.g., GPT-4 or Claude) where access to internal representational trajectories is restricted. 

Second, although our approach successfully bypasses the computational burden of language-specific fine-tuning, extracting and processing full-depth hidden states across multiple layers still incurs non-trivial memory overhead and requires a complete forward pass. This makes it more computationally demanding than simple surface-level heuristics. 

Finally, our multilingual evaluation leverages datasets translated from English using Gemini 2.5 Flash. Despite implementing a rigorous human quality assurance and back-translation pipeline to ensure semantic integrity, evaluating native, naturally occurring prompts in medium and low-resource languages (like Bangla and Amharic) might reveal cultural and linguistic nuances that are not fully captured by translated benchmarks.


\bibliography{custom}

\newpage
\clearpage
\onecolumn

\appendix

\section{Multilingual Data Examples}
\label{sec:appendix_data_examples}

\begin{figure*}[h]
    \centering
    \includegraphics[width=\textwidth]{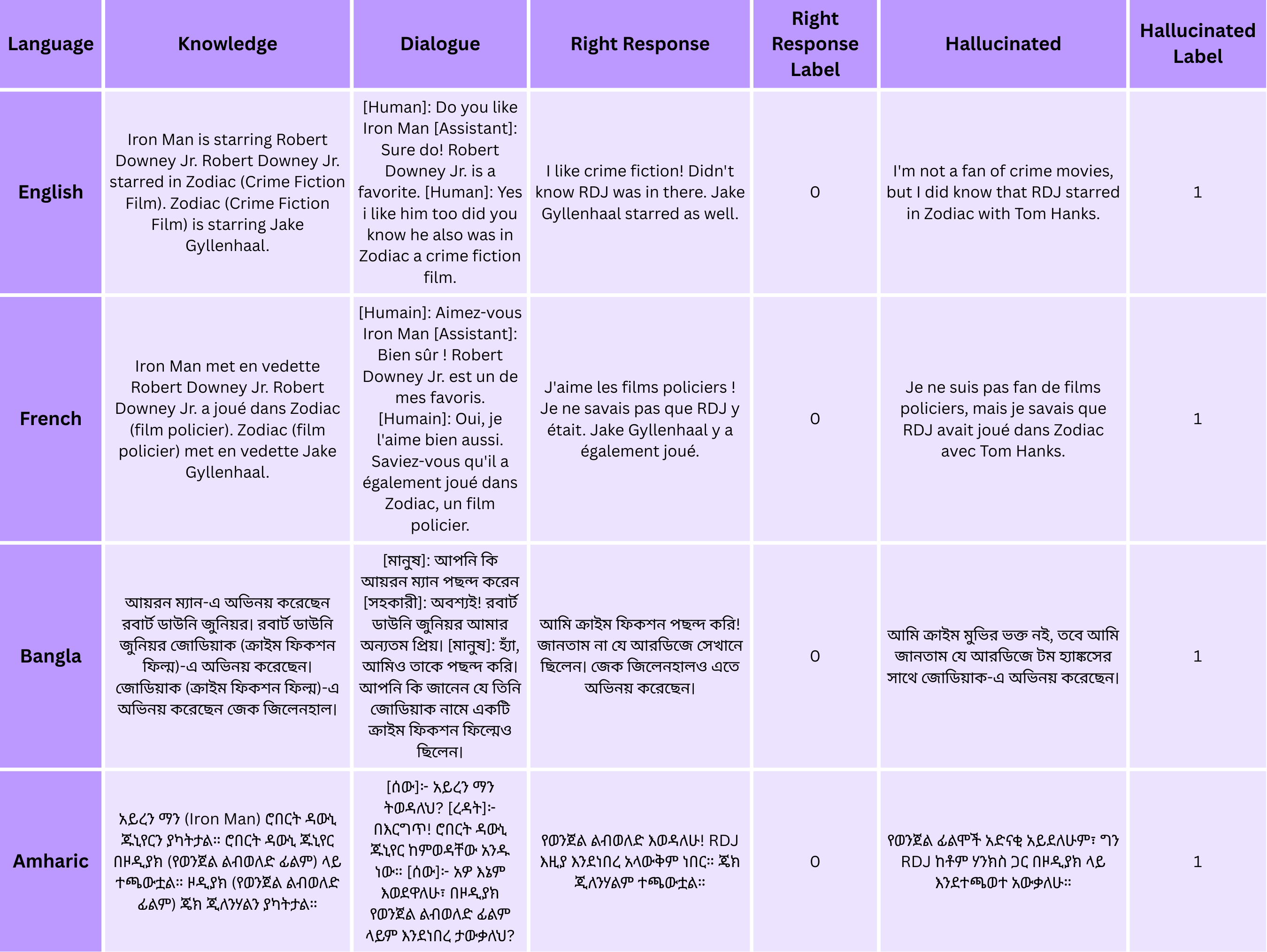}
    \caption{Representative multilingual data points from the HaluEval dataset, demonstrating the semantic preservation of entities and factual inconsistencies across translations.}
    \label{tab:halueval_example}
\end{figure*}

In this section, we provide representative data points from our evaluation benchmarks across all four languages: English (Base), French (High-Resource), Bangla (Medium-Resource), and Amharic (Low-Resource). Table \ref{tab:halueval_example} presents an instance from the HaluEval dataset, illustrating the grounding knowledge, the dialogue history, and both the faithful and hallucinated responses. 

\begin{figure*}[h]
    \centering
    \includegraphics[width=\textwidth]{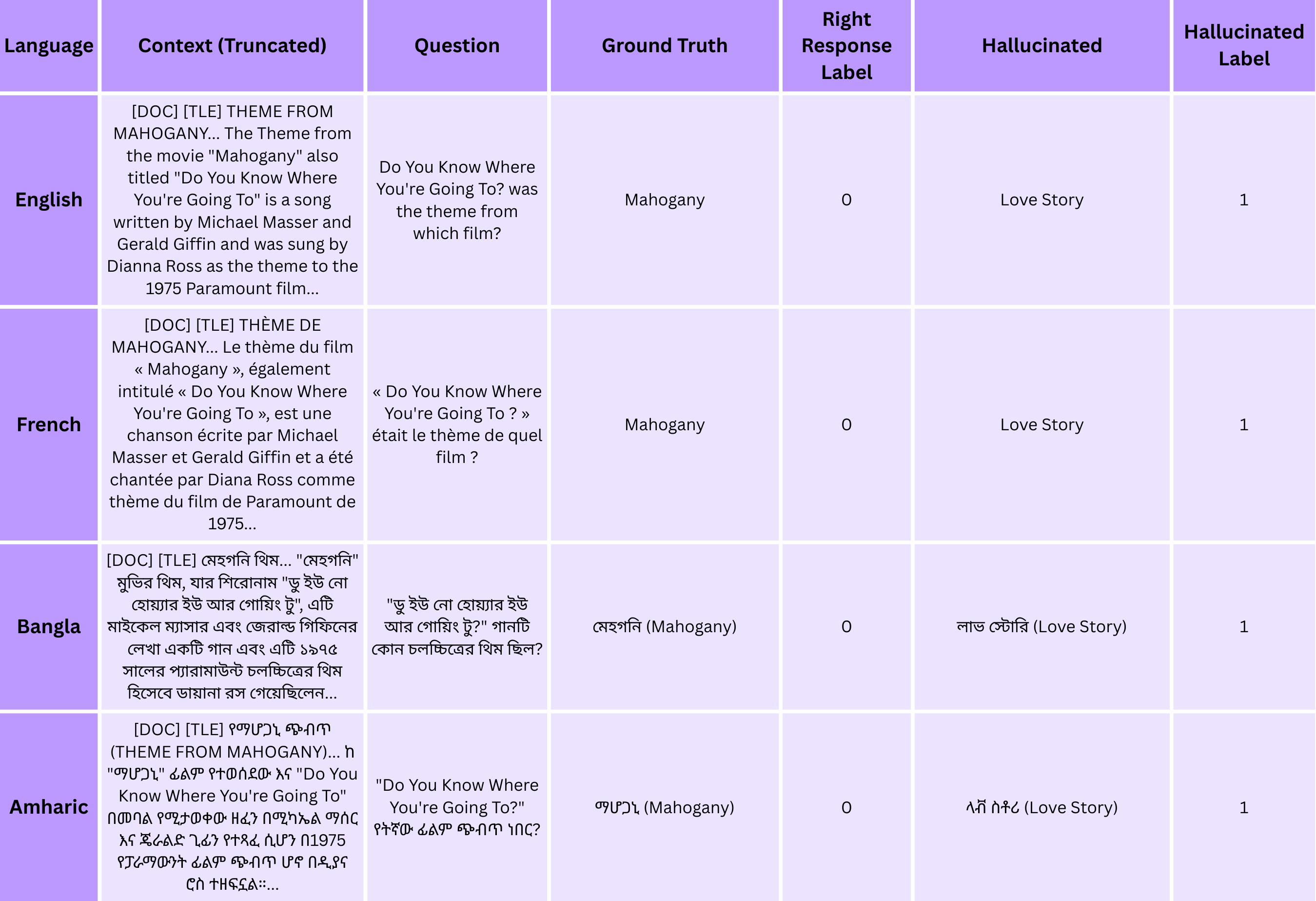}
    \caption{Representative multilingual data points from the modified TriviaQA dataset. The context has been truncated for brevity. The hallucinated response ($a_i^-$) is generated by prompting Gemma-2-B to produce a plausible but factually incorrect answer.}
    \label{tab:triviaqa_example}
\end{figure*}

Table \ref{tab:triviaqa_example} illustrates an example from our modified TriviaQA dataset. TriviaQA \citep{joshi2017triviaqa} is adapted for our evaluation by collecting realistic, model-generated hallucinations. In the original dataset, each entry consists of a question and its ground-truth correct answer. To generate plausible hard negatives, we prompt an early-generation language model known for its propensity to hallucinate, Gemma-2-2B, to answer each question. Responses that are plausible but factually incorrect are extracted to serve as the hallucinated answers. We then construct our final evaluation dataset by pairing each question with its true correct answer and its corresponding model-generated hallucination.

\end{document}